\documentclass[sigconf]{acmart}

\usepackage{amsmath}
\usepackage{lipsum}  
\usepackage{adjustbox}
\usepackage{booktabs}
\usepackage{multicol}
\usepackage{multirow}

\let\emptyset\varnothing
\usepackage{algorithm}
\usepackage{enumitem}
\usepackage[noend]{algpseudocode}

\usepackage{graphicx}
\usepackage{subfig}

\newcommand{\gray}[1]{{\color{gray}{#1}} \addcontentsline{tdo}{todo}{#1} }

\algnewcommand\algorithmicforeach{\textbf{for each}}
\algdef{S}[FOR]{ForEach}[1]{\algorithmicforeach\ #1\ \algorithmicdo}

\AtBeginDocument{%
  }

\copyrightyear{2022} 
\acmYear{2022} 
\setcopyright{acmcopyright}\acmConference[KDD '22]{Proceedings of the 28th ACM SIGKDD Conference on Knowledge Discovery and Data Mining}{August 14--18, 2022}{Washington, DC, USA}
\acmBooktitle{Proceedings of the 28th ACM SIGKDD Conference on Knowledge Discovery and Data Mining (KDD '22), August 14--18, 2022, Washington, DC, USA}
\acmPrice{15.00}
\acmDOI{10.1145/3534678.3539476}
\acmISBN{978-1-4503-9385-0/22/08}

\begin{document}

\title{In Defense of Core-set:\\A Density-aware Core-set Selection for Active Learning }

\author{Yeachan Kim}
\affiliation{%
  \institution{Deargen Inc.}
  \city{Seoul}
  \country{South Korea}}
\email{yeachan@deargen.me}

\author{Bonggun Shin}
\affiliation{%
  \institution{Deargen USA, Inc.}
  \city{Atlanta}
  \country{GA}}
\email{bonggun.shin@deargen.me}

\renewcommand{\shortauthors}{Yeachan Kim \& Bonggun Shin}

\begin{abstract}
Active learning enables the efficient construction of a labeled dataset by labeling informative samples from an unlabeled dataset. In a real-world active learning scenario, considering the diversity of the selected samples is crucial because many redundant or highly similar samples exist. Core-set approach is the promising diversity-based method selecting diverse samples based on the distance between samples. However, the approach poorly performs compared to the uncertainty-based approaches that select the most difficult samples where neural models reveal low confidence. In this work, we analyze the feature space through the lens of the density and, interestingly, observe that locally sparse regions tend to have more informative samples than dense regions. Motivated by our analysis, we empower the core-set approach with the density-awareness and propose a density-aware core-set (DACS). The strategy is to estimate the density of the unlabeled samples and select diverse samples mainly from sparse regions. To reduce the computational bottlenecks in estimating the density, we also introduce a new density approximation based on locality-sensitive hashing. Experimental results clearly demonstrate the efficacy of DACS in both classification and regression tasks and specifically show that DACS can produce state-of-the-art performance in a practical scenario. Since DACS is weakly dependent on neural architectures, we present a simple yet effective combination method to show that the existing methods can be beneficially combined with DACS.
\end{abstract}



\maketitle

\section{Introduction}
While deep neural networks (DNNs) have significantly advanced in recent years, collecting labeled datasets, which is the driving force of DNNs, is still laborious and expensive. This is more evident in complex tasks requiring expert knowledge for labeling. Active learning (AL) \cite{cohn1996active} is a powerful technique that can economically construct a dataset. Instead of labeling arbitrary samples, AL seeks to label the specific samples that can lead to the greatest performance improvement. AL has substantially minimized the labeling costs in various fields, such as image processing \cite{wana2021nearest}, NLP \cite{hartford2020exemplar}, recommender systems \cite{cai2019multi}, and robotics \cite{chao2010transparent}.

Recent AL approaches are categorized into two classes: \textit{uncertainty}-based and \textit{diversity}-based approaches. The former literally estimates the uncertainty of the samples through the lens of loss \cite{yoo2019learning}, predictive variance \citep{gal2017deep,kirsch2019batchbald}, and information entropy \cite{joshi2009multi}. However, the selection of duplicate or very similar samples is a well-known weakness of this approach. The latter approach selects diverse samples that can cover the entire feature space by considering the distance between samples \cite{sener2017active,agarwal2020contextual}. Although this approach can sidestep the selection of duplicate samples by pursuing diversity, it can be suboptimal due to the unawareness of the informativeness of the selected samples.


Core-set \cite{sener2017active} is one of the most promising approaches in diversity-based methods. It selects diverse samples so that a model trained on the selected samples can achieve performance gains that are competitive with that of a model trained on the remaining data points. The importance of the method can be found in a real-world scenario where there are plenty of redundant or highly similar samples. However, the core-set approach often poorly performs compared to the uncertainty-based methods. One susceptible factor is the selection area over the feature space because the core-set equally treats all samples even though each unlabeled sample has different levels of importance and influence when used to train a model \cite{ren2020not}.

\begin{figure*}[t]
\centering
\subfloat[Entropy vs. Density]{%
  \includegraphics[width=0.24\textwidth]{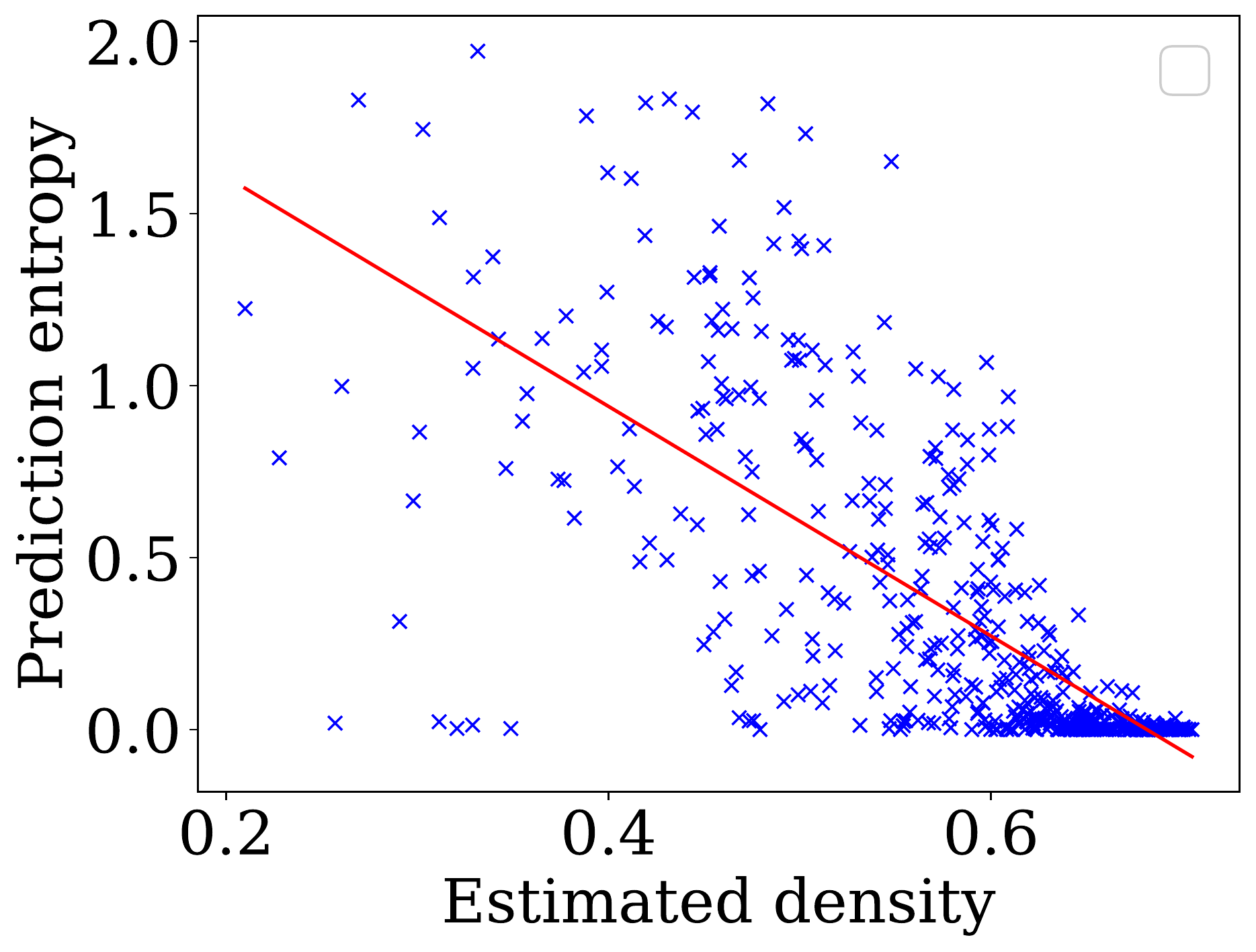}%
  \label{subfig:ent}
}
\subfloat[Loss vs. Density]{%
  \includegraphics[width=0.24\textwidth]{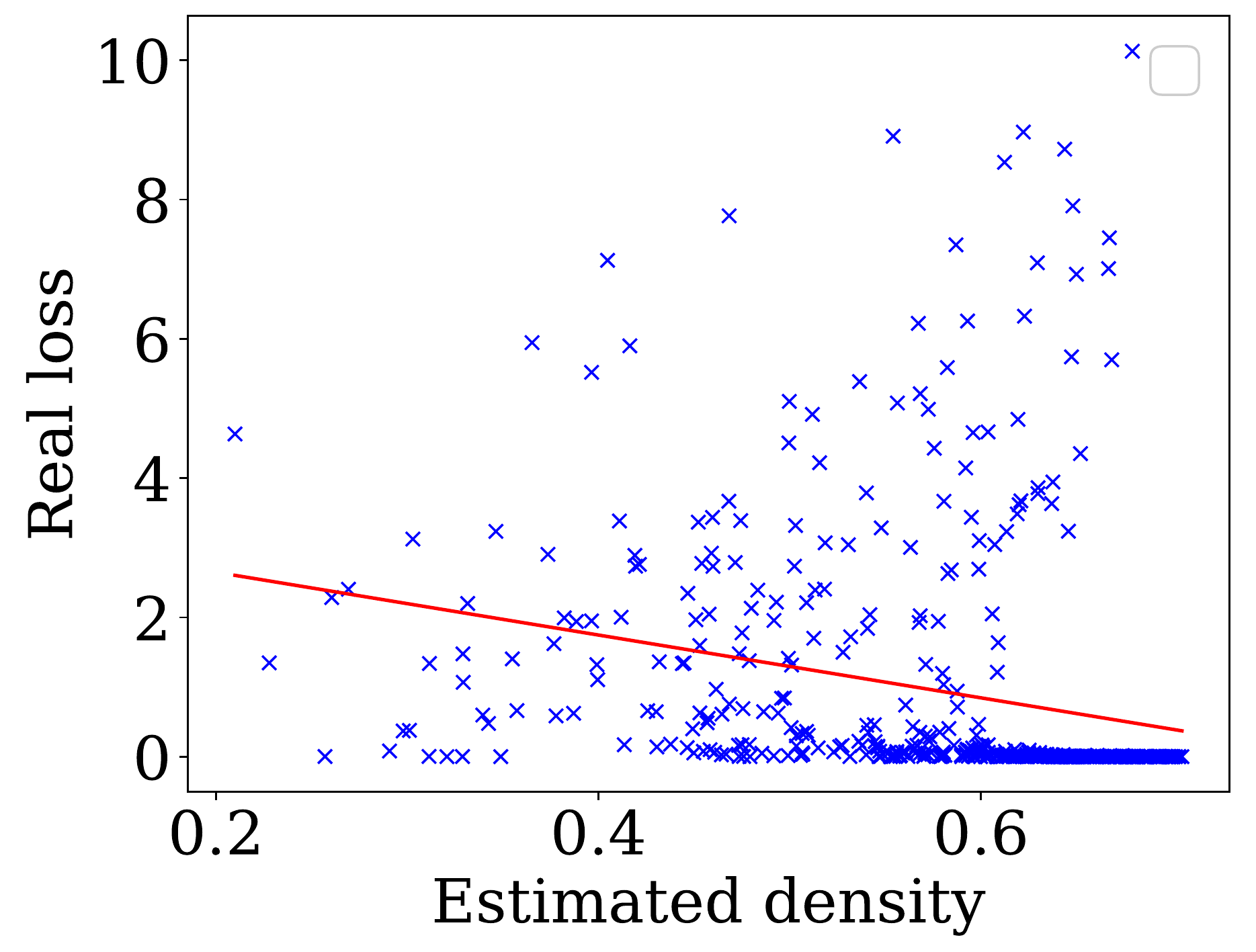}%
  \label{subfig:loss}
}
\subfloat[Label map]{%
  \includegraphics[width=0.25\textwidth]{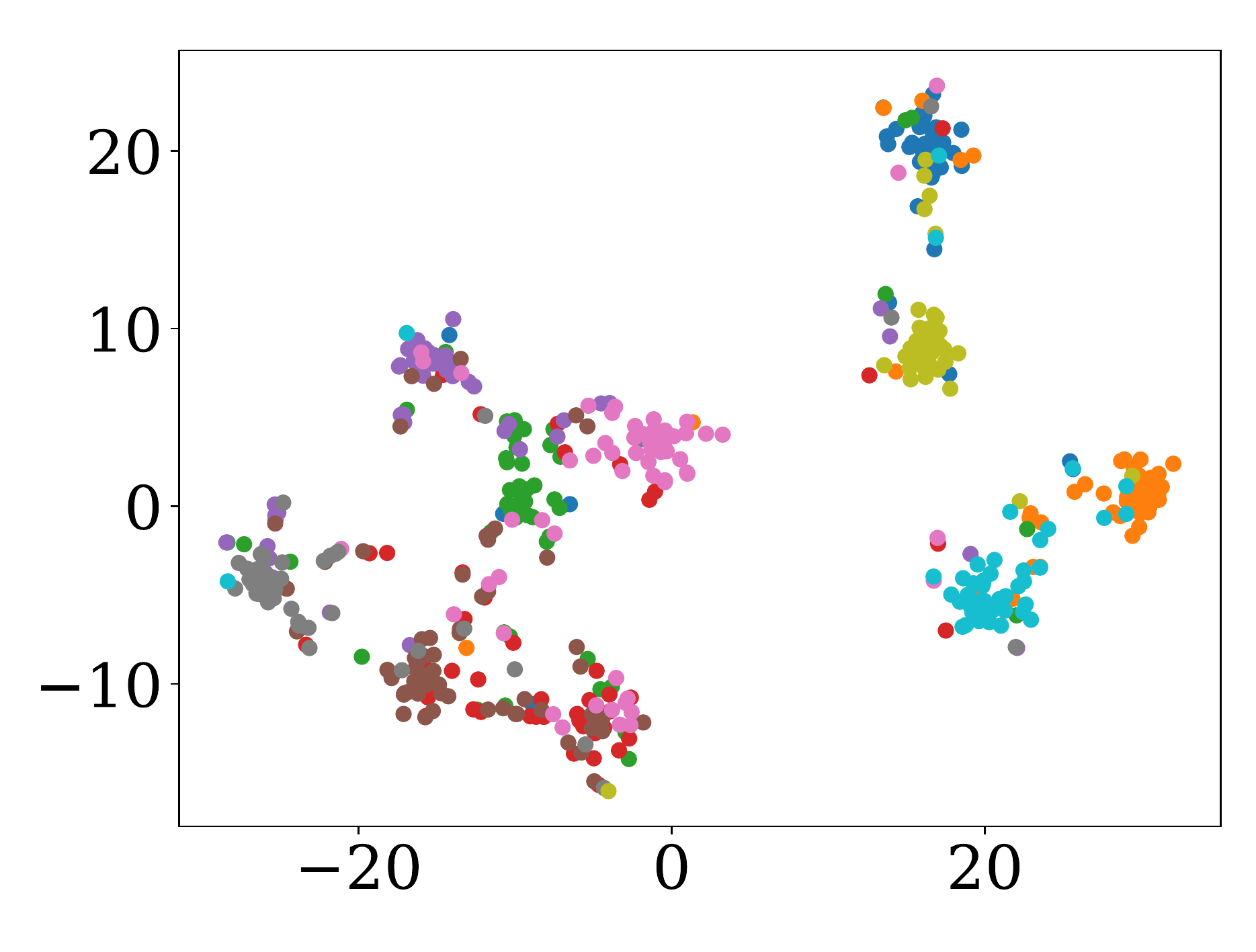}%
  \label{subfig:class}
}
\subfloat[Density map]{%
  \includegraphics[width=0.25\linewidth]{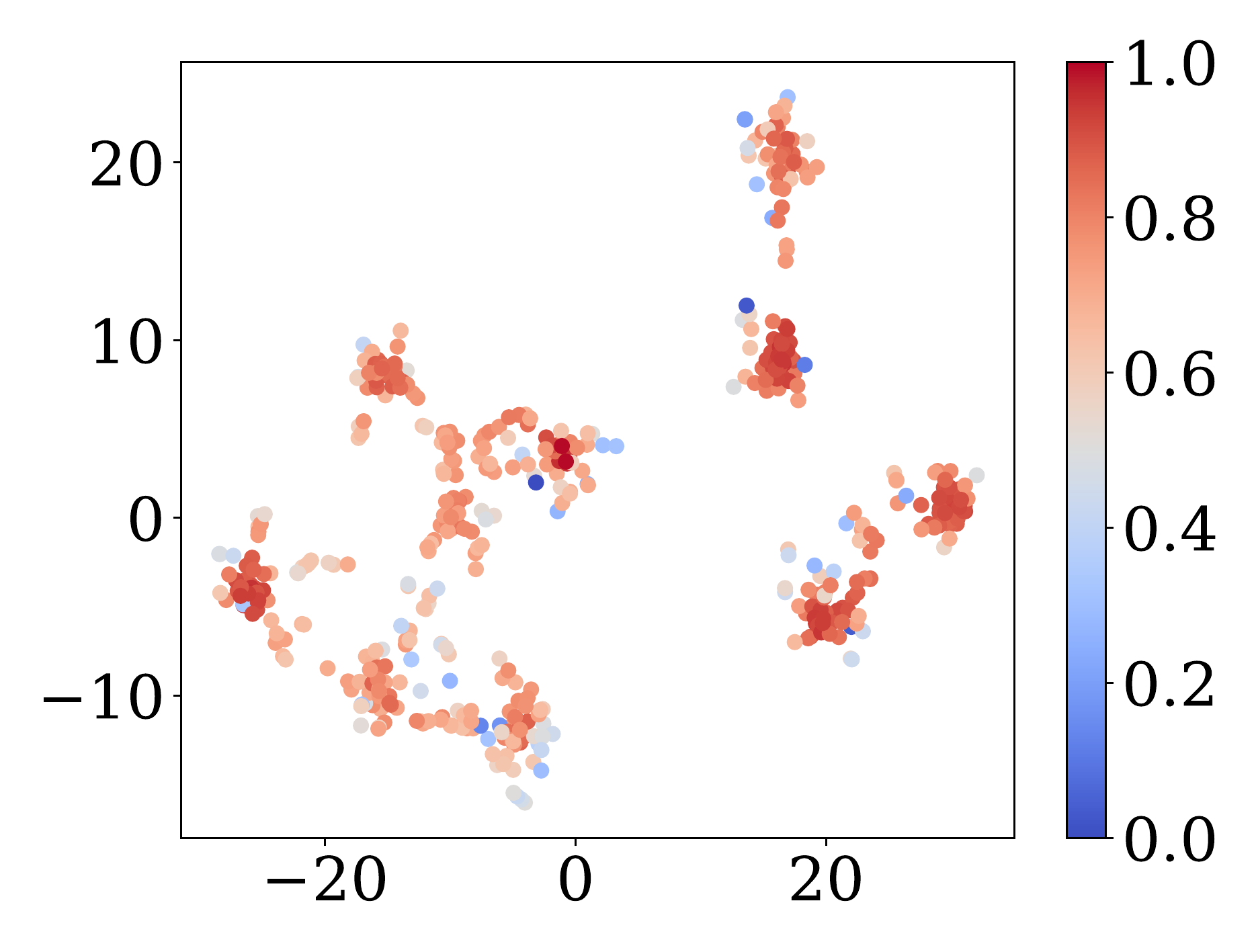}%
  \label{subfig:density}
}
\caption{Correlation plots between density and information entropy ($\rho$ = -0.71), density and loss ($\rho$ = -0.27), and 2-d projected samples with its label and density. In the density map, a value of near one indicates dense regions (Red), and a value of near zero means sparse regions (Blue). It shows that samples in sparse regions are more informative (i.e., high entropy, high loss) than the samples in dense regions. Best viewed in color.}
\label{density}
\end{figure*}

In this work, we analyze the feature space of neural models through the lens of the local density and informativeness (i.e., information entropy, model loss). Interestingly, we find that samples in locally sparse regions are highly uncertain compared to samples in dense regions. Based on this analysis, we propose a density-aware core-set (DACS) which estimates the local density of the samples and selects the diverse samples mainly from the sparse regions. Unfortunately, estimating the density for all samples can lead to computational bottlenecks due to the high dimensionality of feature vectors and a large number of unlabeled samples. To circumvent these bottlenecks, we introduce a density approximation based on locality-sensitive hashing \cite{rajaraman2011mining} to the features obtained from a low-dimensional auxiliary classifier. Note that DACS is task-agnostic and weakly dependent on neural network architecture, revealing that DACS can be favorably combined with any uncertainty-based methods. We thus present a simple yet effective combination method to encourage existing methods to benefit from our work.

We evaluate the effectiveness and the general applicability of DACS on both a classification task (image classification) and a regression task (drug and protein interaction). Comprehensive results and in-depth analysis demonstrate our hypothesis that sampling from the sparse regions is strongly contributed to the superior performance. Moreover, we show that DACS can consistently reach a stable and strong performance in a simulated real-world scenario where highly similar samples exist. In summary, our major contributions include followings:
\begin{itemize}[leftmargin=*]
    \item We propose a novel density-aware core-set method for AL with the analysis of the feature space, which has a novel viewpoint to the diversity-based approach. To circumvent computational bottlenecks, we also propose a new density approximation method.
    
    \item We introduce an effective method for combining DACS with other uncertainty-based methods. Once combined, DACS can work synergistically with other methods, resulting in substantially improved performance.
    
    \item The proposed method significantly improves the performance of the core-set and outperforms strong baselines in both classification and regression tasks. Surprisingly, we also find that DACS selects informative samples fairly well when compared with uncertainty-based methods, even though informativeness is not explicitly considered.
\end{itemize}

\section{Problem Setup for Active Learning}\label{sect:ps}
The objective of active learning is to obtain a certain level of prediction accuracy with the least amount of budgets for constructing a dataset. The setup consists of the unlabeled dataset $(x_i,) \in \mathcal{U}$, the labeled dataset $(x_i, y_i) \in \mathcal{L}$, and the neural model $\mathcal{M}$ parameterized by $\omega$. In the image classification case, $x_i$ and $y_i$ are the input image and its corresponding label, respectively. We define an acquisition function $\varphi(\cdot)$ that returns the most informative or diverse samples within the limited budget as follows:
\begin{equation}\label{acq}
    \mathcal{S} = \{x_1, x_2, ..., x_b\} = \varphi(\mathcal{U}; \mathcal{M}, b)
\end{equation}
where $\mathcal{S}$ is the selected subset with the query budget $b$. After querying the subset to an oracle for its label, we continue to train the model $\mathcal{M}$ on the combined labeled dataset (i.e., $\mathcal{L} \leftarrow \mathcal{L} \cup \mathcal{S}$). The above process is cyclically performed until the query budget is exhausted. To denote each cycle, we add a subscript $t$ to both labeled and unlabeled datasets. For example, the initial labeled and unlabeled datasets are $\mathcal{L}_0$ and $\mathcal{U}_0$, respectively, and the datasets after $t$ cycles are denoted as $\mathcal{L}_t$ and $\mathcal{U}_t$.

\begin{figure*}[t]
    \centering
    \includegraphics[width=0.8\textwidth]{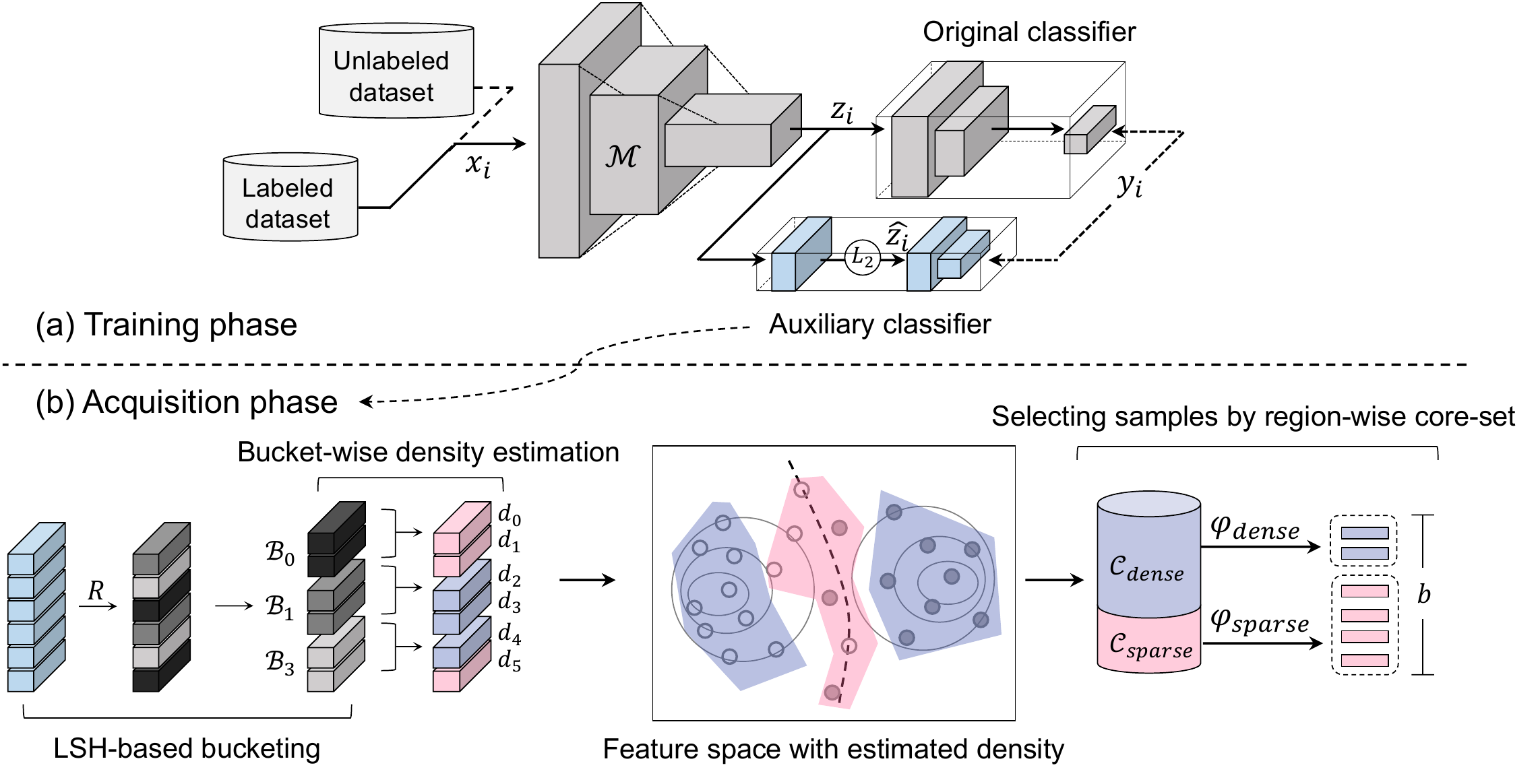}
    \caption{Conceptual overview of density-aware core-set selection. In the training phase, we have the auxiliary classifier that handles low-dimensional and normalized features compared to existing networks (e.g., ResNet). In the acquisition phase, these vectors are used to efficiently estimate the density of unlabeled samples with locality-sensitive hashing.}
    \label{model}
\end{figure*}

\section{Uncertain Regions on Feature Space}\label{sect:observation}

The core-set approach selects diverse samples over the entire feature space (i.e., all unlabeled samples) even though each sample has a different level of influence on the training \cite{ren2020not}. Therefore, if the core-set method can be aware of the informative regions, the method could achieve both uncertainty and diversity at the same time. To this end, we characterize the feature space through the lens of the local density and analyze which density regions are closely related to the informativeness. We quantify the informativeness of unlabeled samples as prediction entropy and loss, which are the popular uncertainty measures, and the density is estimated by averaging the distance between the 20 nearest-neighbor samples' features. 

Figure \ref{density} presents the correlation plots between the estimated density and the uncertainty measures and 2-d visualization of the feature vectors with their density\footnote{Here, we train ResNet-18 \cite{he2016deep} on randomly selected 3,000 samples from CIFAR-10 datasets and visualize the features of unlabeled samples using t-SNE \cite{van2008visualizing}. The settings are detailed in Section \ref{image-model}}. As can be seen from the correlation plots, the density has a negative correlation with the uncertainty measures, and its negative correlation with information entropy is especially strong. In other words, the samples in sparse regions tend to have more information than the samples in dense regions. We also observe that samples in the highly dense regions (Figure \ref{subfig:density}) are clustered well by their labels (Figure \ref{subfig:class}) and, by contrast, the sparse regions include a number of samples that are confusing to the classifier (i.e., not clustered and mixed with other labels). A comprehensive analysis shows that the sparse regions are more informative and uncertain, suggesting that the acquisition should be focused on the locally sparse regions.

The superiority of the sparse region can be explained to some extent by the cluster assumption. Under the assumption that states the decision boundary lies in low density regions \cite{seeger2000learning,chapelle2005semi}, samples in sparse regions can be treated as being near the decision boundary. The near samples of the boundary have high prediction entropy and loss \cite{fawzi2018empirical,karimi2020decision}, which is similar property to samples in sparse regions, indicating that the sparse regions are closely related to the decision boundary. Furthermore, by following the above assumption, samples in dense regions can be regarded as samples in close vicinity to a cluster where the neural models reveal low entropy. This suggests that selecting samples from sparse regions is more effective than selecting samples from dense regions when constructing the dataset.



\section{Density-aware Core-set \\ for Diversity-based Active Learning}



This section details the proposed method, coined density-aware core-set (DACS), that enables the core-set approach to select diverse but informative samples. DACS begins by estimating the local density of the unlabeled samples (Section~\ref{sect:density_estimation}). Afterward, DACS selects diverse samples from the density-estimated regions such that the samples in the sparse regions are mainly selected (Section~\ref{coreset}).


\subsection{Density Estimation for Unlabeled Samples}\label{sect:density_estimation}
\subsubsection{Nearest-neighbor-based density estimation}\label{knn}
\hfill \break The naive method to estimate the local density is to use $k$-nearest neighbors. In this method, the density is simply calculated by the average distance to nearest k samples.
\begin{equation}\label{knn}
d(x_i, k) = \frac{1}{k}\sum_{j \in NN(x_i, k)} \delta(x_i, x_j)
\end{equation}
where $NN(x_{i}, k)$ is the function that returns $k$ nearest samples from the sample $x_i$, $\delta$ is the distance measure (e.g., euclidean distance, angle distance) between two samples that are typically represented as intermediate features \cite{sener2017active}.

However, there are two major computational bottlenecks in $k$-$NN$-based density estimation. The first bottleneck is the large number of unlabeled samples in active learning. To estimate the density of each sample, $NN(\cdot)$ should calculate the distance $\delta$ to all unlabeled samples. The second factor is the high dimensionality of the features of each sample in neural networks, which influences the distance calculation $\delta$ between samples.

\subsubsection{Efficient density estimation}
\hfill \break To circumvent the above computational bottlenecks in estimating the density, we introduce the hashing-based estimation with the auxiliary training in which the low-dimensional vectors are trained to be compatible to the high-dimensional ones.

\noindent
\textbf{Auxiliary Training.}\label{norm}
To sidestep using the high-dimensional vectors, we carefully add an auxiliary classifier to the existing architectures. The classifier has two distinct properties compared to the existing classifier. First, it consists of the low-dimensional layers to the extent that it does not hurt the accuracy. Second, the feature vectors are normalized during the training to encourage the vectors to be more separable and compact in the feature space \cite{aytekin2018clustering}.

The auxiliary classifier takes the large features of the existing network as input. Then, the input vectors are transformed to the low-dimensional normalized vectors.
\begin{equation}\label{normalization}   
    \hat{z_i} = \frac{W_n^{\intercal}z_{i}}{||W_n^{\intercal}z_{i}||_2^2}
\end{equation}
where $z_i$ is the large features of the existing networks, $W_n \in \mathbb{R}^{d \times d_n}$ is a learnable weight matrix, $d$ and $d_n$ are the dimensionality of original and normalized vectors, respectively, and $d_n \ll d$. From the viewpoint of the large feature vector $z_i$, the loss function in the classification case is defined as:
\begin{equation}\label{cross-entropy}
    \mathfrak{L}(z_i;\omega_t) = -\frac{1}{|\mathcal{L}_t|}\sum_{(x_i, y_i) \in \mathcal{L}_t} y_i \cdot log P(z_i;\omega_t)
\end{equation}
where $y$ is the ground-truth label and $P$ is the predicted probability given feature vectors $z_i$ and the model parameters $\omega_t$. The overall loss function with the auxiliary training can be represented as follows:
\begin{equation}\label{objective}
    \mathfrak{L}_{total} = \mathfrak{L}(z_i;\omega_t) + \lambda \cdot \mathfrak{L}(\hat{z_i};\omega_t \cup \hat{\omega_t})
\end{equation}
where $\lambda$ is the control factor of the normalized training, and $\hat{\omega_t}$ is the additional parameters of the auxiliary classifier. As the training with auxiliary classifier might hurt the performance of the main classifier, we prevent the gradient flow between the main and auxiliary classifier after specified training epochs (see Section~\ref{image-model} for more information). In the acquisition phase, we use the low-dimensional normalized vectors instead of large features.

\noindent
\textbf{Hashing-based Density Estimation}\label{attention}
Auxiliary training results in computationally efficient and well-separated vectors for unlabeled samples. However, the large number of unlabeled samples is still a potential bottleneck to find nearest neighbors (i.e., $NN(\cdot)$ in Eq.~\eqref{knn}), which is the necessary process to estimate the density. 

Locality-sensitive hashing (LSH) has been adopted to address the computational bottlenecks of neural networks \cite{kitaev2020reformer,chen2020mongoose}. The LSH scheme assigns the nearby vectors into the same hash with high probability. To reduce the bottleneck of a large number of samples in estimating the density, the samples are hashed into $k$ different buckets, and finding nearest neighbors in each bucket instead of the entire dataset enables the efficient estimation for the density. To obtain $k$ buckets, we apply a fixed random rotation $R \in \mathbb{R}^{d_n \times \frac{k}{2}}$ to the feature vectors and define the hashed bucket of $x_i$ as follows:
\begin{equation}\label{rot}
    s_i = \text{arg max} ([R^\intercal \hat{z_i};-R^\intercal \hat{z_i}])
\end{equation}
where $[u; v]$ is the concatenation of two vectors. $s_i$ indicates the bucket number of the sample $x_i$.

The above hashing process assigns a different number of samples to each bucket, preventing batch computation. For batch processing, the samples are sorted by corresponding bucket numbers, and the sorted samples are then sequentially grouped by the same size. Formally, the bucket containing $i$-th samples in the sorted order is defined as follows:
\begin{equation}
    \mathcal{B}(i) = \{j \text{ | } \lfloor\frac{i}{m}\rfloor - 1 \leq \lfloor\frac{j}{m}\rfloor \leq \lfloor\frac{i}{m}\rfloor\}
\end{equation}
where $m$ is the size of buckets (i.e., $m = \frac{|\mathcal{U}_t|}{k}$). Within each bucket, the density is estimated by calculating the weighted cosine similarity as follows:
\begin{equation}\label{cosine}
\begin{split}
d_i &= \sum_{j \in \mathcal{B}(i)/i} \sigma(cos\theta_{ij}) \cdot cos\theta_{ij}  \\
 &= \sum_{j \in \mathcal{B}(i)/i} \sigma(\hat{z_i}^\intercal \hat{z}_j) \cdot \hat{z_i}^\intercal \hat{z}_j \text{  ($ \because ||\hat{z}_i|| = ||\hat{z}_j|| = 1$)}\\
\end{split}
\end{equation}
where $\sigma(\cdot)$ is the sigmoid function, and $\theta_{ij}$ is the angle between $\hat{z}_i$ and $\hat{z}_j$. To favor near samples while reducing the effect of distant samples, sigmoid weights are applied to the similarity. Since the sizes of all buckets are the same as $\lfloor{m}\rfloor$, Eq. \ref{cosine} can be viewed as calculating the similarity between fixed $\lfloor{m}\rfloor$-nearest neighbor samples, and the estimates are comparable across different buckets. This naturally makes the samples in the dense region have higher estimates than that of the sparse region because the samples in the dense have the more close samples in each bucket.

\begin{algorithm} \caption{Density-aware Core-set} \label{algo}
\begin{algorithmic}[1]
\Require labeled pool $\mathcal{L}_t$, unlabeled pool $\mathcal{U}_t$, density group $\mathcal{C}$, query budget $b$
\State $\mathcal{S} \leftarrow \emptyset$
\State \gray{\# Perform core-set selection in each group $\mathcal{C}$ }
\ForEach {$\mathcal{C}_i \in \mathcal{C} $} 
\State \gray{\# Calculate the selection ratio from Eq. \ref{sample_num}}
\State $n_{i} \leftarrow \lfloor r_i \cdot b \rfloor \text{ and } S_i \leftarrow \emptyset$ 
\State \gray{\# Find $n_i$ center points in a greedy manner}
\Repeat 
\State $u \leftarrow \text{arg }\min_{j \in \mathcal{C}_i/S_i}\max_{m \in (\mathcal{L}_t \cup S_i) }\text{ }cos\theta_{jm}$
\State $S_i \leftarrow S_i \cup u$
\Until $|S_i| < n_i$
\State $\mathcal{S} \leftarrow \mathcal{S} \cup S_i$
\EndFor
\State \gray{\# Update unlabeled and labeled dataset}
\State $\mathcal{L}_{t+1} \leftarrow \mathcal{L}_t \cup \mathcal{S}$
\State $\mathcal{U}_{t+1} \leftarrow \mathcal{U}_t \cap {\mathcal{S}}^c$
\end{algorithmic}
\end{algorithm}

\subsection{Density-aware Core-set Selection}\label{coreset}

\begin{table*}[t]
\caption{Active learning results on CIFAR-10. Evaluation metric is test accuracy (\%). Best and second best results are highlighted in \textbf{boldface} and \underline{underlined}, respectively. With the full dataset, we achieve 92.1\% accuracy on the test dataset.}
\begin{adjustbox}{max width=\textwidth}
\begin{tabular}{@{}lcccccccccc@{}}
\toprule
                                              & \multicolumn{10}{c}{Proportion of Labeled Samples (CIFAR-10)}              \\ \cmidrule(l){2-11} 
Methods                                       & 2\% & 4\% & 6\% & 8\% & 10\% & 12\% & 14\% & 16\% & 18\% & 20\% \\ \midrule
Random                                          & 50.6 & 59.9 & 67.2 & 74.5 & 79.5 & 83.3 & 84.8 & 85.2 & 86.3 & 87.0 \\ \midrule
LearnLoss \cite{yoo2019learning}                & 54.1 & 66.2 & 76.3 & 80.5 & 82.2 & 85.8 & 87.4 & 88.4 & 89.1 & 89.9 \\
NCE-Net \cite{wana2021nearest}                  & \textbf{56.4} & 67.2 & 75.1 & 80.9 & 82.4 & 85.0 & 87.1 & 88.5 & 90.2 & 90.2 \\
CDAL \cite{agarwal2020contextual}               & 50.6 & 64.1 & 73.7 & 80.2 & 83.8 & 86.1 & 87.3 & 88.6 & 89.5 & 90.2 \\
Core-set \cite{sener2017active}                 & 50.6 & 62.4 & 71.1 & 78.2 & 82.2 & 84.5 & 86.6 & 88.1 & 89.1 & 89.8 \\
DACS (Ours.)                                    & 50.6 & \textbf{70.8} & \textbf{79.1} & \underline{82.9} & \underline{84.3} & \underline{86.9} & \underline{88.2} & 89.1 & 89.6 & 90.4 \\ \midrule \midrule
LearnLoss \cite{yoo2019learning} + DACS (Ours.)  & 53.0 & 69.9 & 78.5 & \textbf{83.2} & 84.2 & \textbf{87.2} & \textbf{88.6} & \textbf{89.6} & \textbf{90.6} & \textbf{91.2} \\
NCE-Net \cite{wana2021nearest} + DACS (Ours.)    & \underline{55.2} & \underline{70.6} & \underline{78.9} & 82.5 & \textbf{84.6} & 86.3 & 88.2 & \underline{89.4} & \underline{90.3} & \underline{91.1} \\\bottomrule
\end{tabular}
\end{adjustbox}
\label{exp:CIFAR-10}
\end{table*}

Based on the efficiently estimated density, we select core-set samples in which the samples in the sparse regions are more favorably selected. To this end, we first divide the unlabeled samples into dense or sparse regions by performing Jenks natural breaks optimization\footnote{Jenks natural breaks optimization is analogous to the single dimensional k-means clustering. The objective is to determine which set of breaks leads to the smallest intra-cluster variance while maximizing inter-cluster variance.} \cite{jenks1967data} to the estimated densities, resulting in $h$ different groups which are clustered by the density, and these groups are denoted as follows:
\begin{equation}\label{cluster}
    \mathcal{C} = \{\mathcal{C}_1, \mathcal{C}_2, ... , \mathcal{C}_h\} \text{ where } \mathcal{C}_i = \{x^{i}_1, x^{i}_2, ..., x^{i}_{|\mathcal{C}_i|}\}
\end{equation}
where $x^i_{j}$ is the $j$-th sample in the cluster $\mathcal{C}_i$. Over the different density clusters, we perform k-center greedy \cite{wolf2011facility} to select diverse sample. As the entire feature space is divided into $h$ regions (i.e., from dense to sparse regions), the k-center selection in the core-set \cite{sener2017active} is also decomposed with the same number of clusters. The number of centers in decomposed greedy selection are determined by inverse proportion of the size of the cluster $\mathcal{C}_i$ because the groups clustered by high density tend to occupy more data than the groups with relatively low density. Such strategy enables to select more samples from the sparse regions and the selection ratio can be defined as:
\begin{equation}\label{sample_num}
    r_i = \text{softmax}((1 - \frac{|\mathcal{C}_i|}{|\mathcal{U}_t|})/\tau)
\end{equation}
where $\tau$ is a temperature that controls the sharpness of the distribution. The detailed process of density-aware core-set is described in Algorithm \ref{algo}. Note that we replace the euclidean distance with the cosine similarity since the feature are normalized in the auxiliary training. The comprehensively selected subset from the method is represented as follows:
\begin{equation}
    \mathcal{S} = S_1 \cup ... \cup S_h \text{ where } S_i = \varphi_i(\mathcal{C}_i; \mathcal{M}, \lfloor r_i \cdot b \rfloor)
\end{equation}
where $\varphi_{i}(\cdot)$ is the core-set-based acquisition function in cluster $\mathcal{C}_i$. After selecting the subset $\mathcal{S}$, we query the subset to the oracle for its labels and perform the next cycle of the active learning on the updated dataset.

\subsection{Combination with Uncertainty-based Selection Methods}\label{sect:comb}
DACS can be complementarily combined with uncertainty-based methods because the proposed method naturally sidesteps the selection of duplicate samples. We take a “\textit{expand and squeeze}” strategy to combine DACS with the uncertainty-based method. Specifically, DACS pre-selects $\eta$ times more samples than the query budget $b$ as query candidates. Then, the uncertainty-based method sorts the candidates by its uncertainty criterion and finally selects the most uncertain $b$ sample. Since DACS selects diverse samples as useful candidates, the uncertainty-based methods are free of selecting redundant or highly similar samples in the acquisition. Furthermore, in the case where DACS may overlook informative samples in the center selection, the uncertainty-based method can correct the missed selection.

\begin{table*}[t]
\caption{Active learning results on RMNIST. Evaluation metric is Top-1 test accuracy (\%). Best and second best results are highlighted in \textbf{boldface} and \underline{underlined}, respectively. With the full dataset, we achieve 98.3\% accuracy on the test dataset.}
\begin{adjustbox}{max width=\textwidth}
\begin{tabular}{@{}lcccccccccc@{}}
\toprule
                                              & \multicolumn{10}{c}{Proportion of Labeled Samples (RMNIST)}              \\ \cmidrule(l){2-11} 
Methods                                         & 0.2\% & 0.4\% & 0.6\% & 0.8\% & 1.0\% & 1.2\% & 1.4\% & 1.6\% & 1.8\% & 2.0\% \\ \midrule
Random                                          & 85.1 & 88.2 & 90.8 & 92.3 & 93.1 & 93.7 & 94.2 & 94.2 & 94.3 & 94.7 \\ \midrule
LearnLoss \cite{yoo2019learning}                & 86.1 & 88.5 & 90.1 & 91.6 & 92.4 & 93.0 & 93.5 & 94.4 & 94.5 & 94.5 \\
NCE-Net \cite{wana2021nearest}                  & \underline{87.5} & 88.9 & 90.5 & 91.9 & 93.3 & 93.8 & 94.7 & 95.1 & 95.6 & 95.6 \\
CDAL \cite{agarwal2020contextual}               & 85.1 & 90.5 & 93.2 & \underline{94.8} & \underline{95.8} & \underline{95.9} & 96.0 & 96.2 & 96.5 & 96.7 \\
Core-set \cite{sener2017active}                 & 85.1 & 88.6 & 91.9 & 93.0 & 94.3 & 94.6 & 95.6 & 96.1 & 96.1 & 96.5 \\
DACS (Ours.)                                    & 85.1 & \textbf{92.1} & \textbf{94.3} & \textbf{95.1} & \textbf{96.0} & \textbf{96.4} & \textbf{96.8} & \textbf{97.0} & \textbf{97.4} & \textbf{97.6} \\ \midrule \midrule
LearnLoss \cite{yoo2019learning} + DACS (Ours.)  & 85.7 & 90.1 & 91.4 & 93.4 & 93.7 & 94.6 & 95.5 & 96.1 & \underline{96.6} & \underline{97.0} \\
NCE-Net \cite{wana2021nearest} + DACS (Ours.)    & \textbf{87.6} & \underline{91.3} & \underline{93.3} & 94.8 & 95.7 & 95.8 & \underline{96.2} & \underline{96.3} & 96.6 & 96.8 \\\bottomrule
\end{tabular}
\end{adjustbox}
\label{exp:rmnist}
\end{table*}

\begin{table*}[t]
\caption{Active learning results on Davis. Evaluation metric is mean squared errors (i.e., lower errors indicates better models). Best and second best results are highlighted in \textbf{boldface} and \underline{underlined}, respectively. With the full dataset, we achieve 0.271 MSE on the test dataset.}
\begin{adjustbox}{max width=\textwidth}
\begin{tabular}{@{}lcccccccccc@{}}
\toprule
                                              & \multicolumn{10}{c}{Proportion of Labeled Samples (Davis)}              \\ \cmidrule(l){2-11} 
Methods                                         & 5\% & 10\% & 15\% & 20\% & 25\% & 30\% & 35\% & 40\% & 45\% & 50\% \\ \midrule
Random                                          & 0.638 & \textbf{0.624} & 0.597 & 0.574 & 0.554 & 0.512 & 0.442 & 0.421 & 0.387 & 0.366 \\ \midrule
LearnLoss \cite{yoo2019learning}                & \textbf{0.636} & 0.631 & 0.617 & 0.571 & 0.514 & 0.478 & 0.421 & 0.376 & 0.345 & 0.331 \\
Core-set \cite{sener2017active}                 & 0.638 & 0.637 & 0.604 & 0.584 & 0.541 & 0.422 & 0.407 & 0.372 & 0.350 & 0.337 \\
DACS (Ours.)                                    & {0.638} & 0.635 & \underline{0.543} & \textbf{0.465} & \textbf{0.423} & \underline{0.397} & \underline{0.361} & \underline{0.332} & \textbf{0.306} & \textbf{0.298} \\ \midrule \midrule
LearnLoss \cite{yoo2019learning} + DACS (Ours.)  & \underline{0.637} & \underline{0.626} & \textbf{0.513} & \underline{0.488} & \underline{0.429} & \textbf{0.385} & \textbf{0.356} & \textbf{0.331} & \underline{0.324} & \underline{0.312} \\ \bottomrule
\end{tabular}
\end{adjustbox}
\label{exp:davis}
\end{table*}

\section{Experiments}
In this section, we evaluate the proposed method in the settings of the active learning. We perform two different tasks, which are image classification and drug-protein interaction, to show the strength of DACS in different domains.

\subsection{Experimental Settings}
\subsubsection{Baselines.} We compare DACS with the four strong baselines which include two uncertainty-based methods ({LearnLoss} \cite{yoo2019learning} and {NCE-Net} \cite{wana2021nearest}) and two diversity-based methods ({Core-set} \cite{sener2017active} and {CDAL} \cite{agarwal2020contextual}). 

\subsubsection{Training configuration.}
For a fair comparison, we perform the comparison on the same settings of an initial labeled set (i.e., $\mathcal{L}_0$) and the same random seeds, and we report the average performance of three trials. For a fair comparison, we not only use the same networks between baselines but also perform auxiliary training for all baselines. We have implemented the proposed method and all experiments with PyTorch \cite{paszke2017automatic} and trained the models on a single NVIDIA Tesla V100 with 32GB of RAM. 

For the hyper-parameters of DACS, the reduced dimension in the auxiliary classifier is set to 16, and we set the number of buckets ($k$ in Eq. \ref{rot}) and the number of breaks ($h$ in Eq. \ref{cluster}) to 100 and 4, respectively. The temperature is set to 0.25 ($\tau$ in Eq. \ref{sample_num}). The above parameters are chosen by validation on CIFAR-10 dataset, and we found that such parameters work fairly well in different datasets in this paper.

\subsection{Image Classification}
\subsubsection{Dataset configuration.}
We use two different datasets for the image classification task. First, we evaluate each method on CIFAR-10 \cite{krizhevsky2009learning} which is the standard dataset for active learning. CIFAR-10 contains 50,000 training and 10,000 test images of 32×32×3 assigned with one of 10 object categories. We also experiment on Repeated MNIST (RMNIST) \cite{kirsch2019batchbald} to evaluate each method in the real-world setting where duplicate or highly similar samples exist. 

RMNIST is constructed by taking the MNIST dataset and replicating each data point in the training set two times (obtaining a training set that is three times larger than the original MNIST). To be specific, after normalizing the dataset, isotropic Gaussian noise is added with a standard deviation of 0.1 to simulate slight differences between the duplicated data points in the training set. RMNIST includes 180,000 training and 10,000 test images of 28×28×1 assigned with one of 10 digit categories. As an evaluation metric, we use classification accuracy. 

The active learning for CIFAR-10 starts with randomly selected 1,000 labeled samples with 49,000 unlabeled samples. In each cycle, each method selects 1,000 samples from unlabeled pool $\mathcal{U}_t$ and adds the selected samples to the current labeled dataset, and this process is repeatedly performed in 10 cycles. For RMNIST, we reduce the size of the initial set and the query budget to 500 samples.

\subsubsection{Models.}\label{image-model}
For CIFAR-10, we use the 18-layer residual network (ResNet-18) \cite{he2016deep}. Since the original network is optimized for the large images (224×224×3), we revise the first convolution layer to extract smaller features by setting the stride and padding of 1 and dropping the max-pooling layer. For RMNIST, we use 3-layered feed-forward networks. We only apply standard augmentation schemes (e.g., horizontal flip and random crop) to the training on CIFAR-10. The other settings are the same between CIFAR-10 and RMNIST. We train each model for 200 epochs with a mini-batch size of 128. We use the SGD optimizer with an initial learning rate of 0.1 and a momentum of 0.4. After 160 epochs, the learning rate is decreased to 0.01. As in \cite{yoo2019learning}, we stop the gradient from the auxiliary classifier propagated to the main classifier after 120 epochs to focus on the main objective.

\subsubsection{Results.}

The evaluation results are shown in Table~\ref{exp:CIFAR-10} (CIFAR-10) and Table~\ref{exp:rmnist} (RMNIST). In CIFAR-10, the diversity-based methods underperform compared to uncertainty-based methods in early cycles (e.g., 2-5 cycles). However, it is noteworthy that even though DACS also belongs to the diversity-based approach, it outperforms other methods by a large margin during the same cycles. It means that DACS can select informative samples fairly well with the smaller amount of query budget. In later cycles as well, DACS shows competitive or better performance than other strong baselines. 

\begin{figure*}[h]
\centering 
\subfloat[CIFAR-10]{%
  \includegraphics[width=0.32\textwidth]{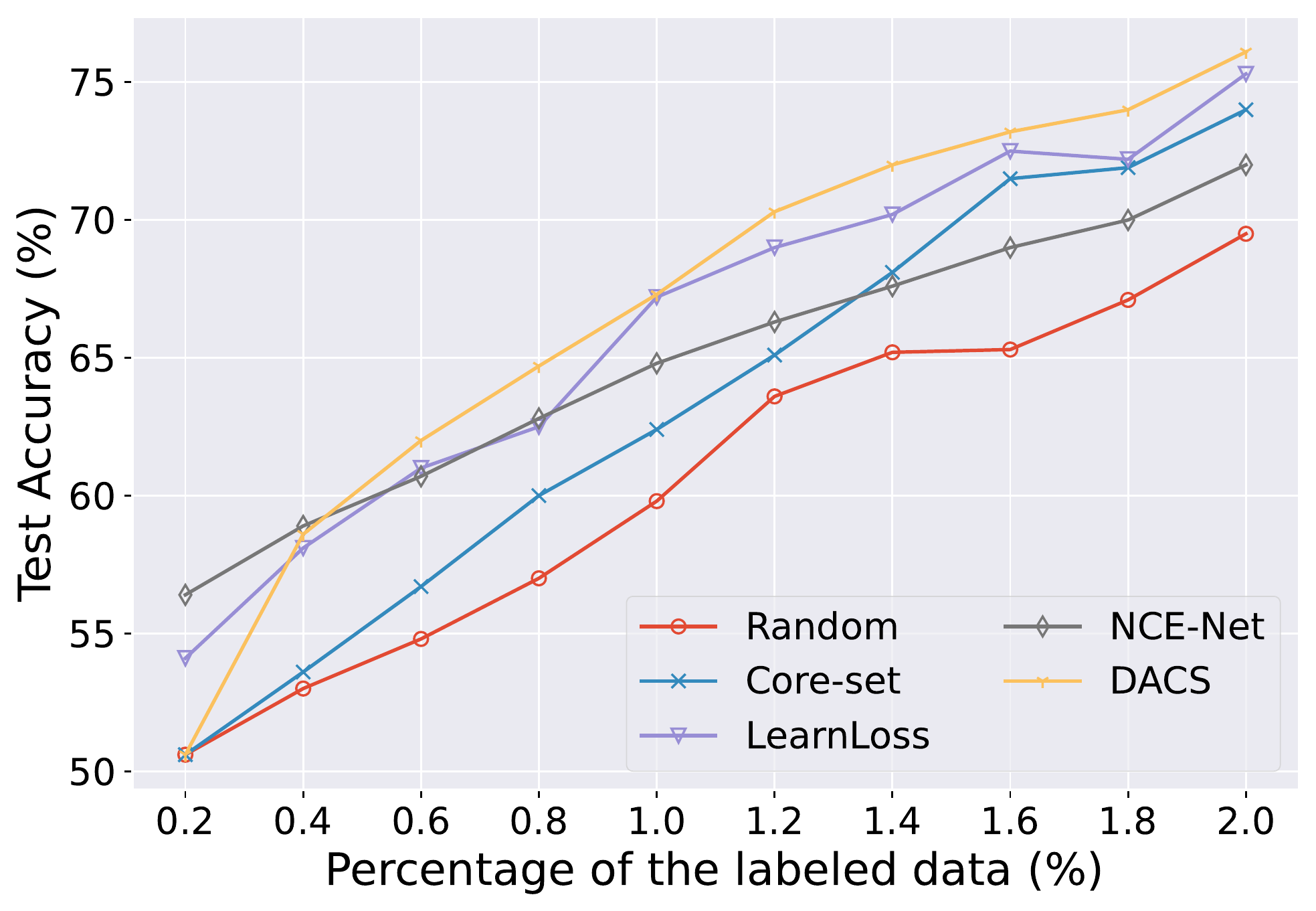}%
  \label{exp:global}
}
\subfloat[RMNIST]{%
  \includegraphics[width=0.32\textwidth]{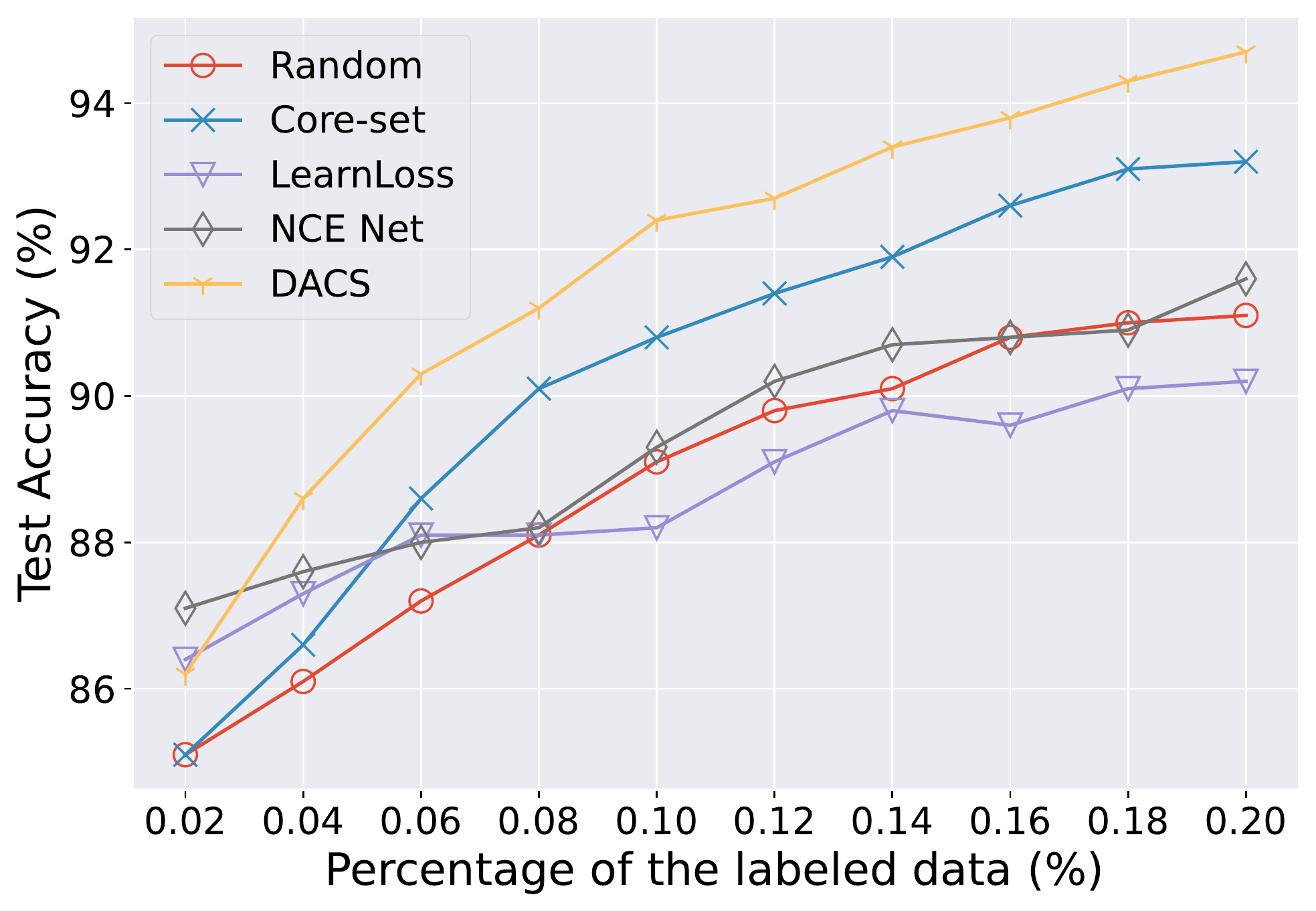}%
  \label{exp:local_drfit}
}
\subfloat[Davis]{%
  \includegraphics[width=0.33\textwidth]{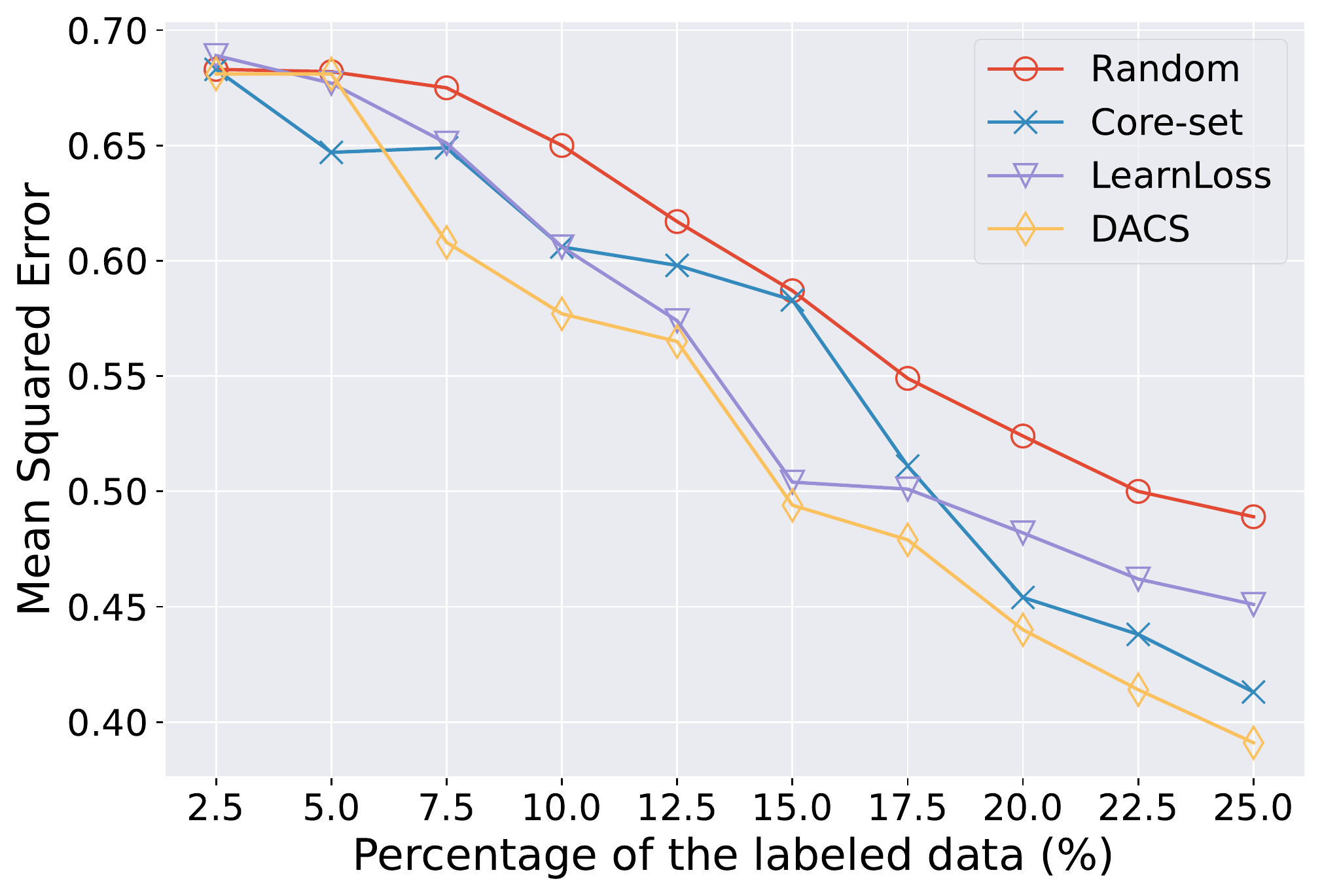}%
  \label{exp:local_drfit}
}
\caption{Active learning results of image classification (a, b) and predicting interaction between drug and target (c). Best viewed in colors.}
\label{fig:small_exp}
\end{figure*}

We can see the strength of DACS in RMNIST which is the more practical scenario. To achieve high accuracy in this dataset, it is significant to consider both diversity and informativeness of samples because redundant samples exist. The uncertainty-based methods poorly perform in the dataset since it mainly considers informativeness and does not aware of the similarity between selected samples. In contrast, the diversity-based methods exhibit their strength over the uncertainty-based method. Particularly, DACS consistently outperforms all baselines in subsequent cycles. For example, DACS better performs on average 2.3\%p and 1.1\%p than uncertainty-based and diversity-based methods, respectively, in the last cycle.

It is noticeable that DACS can be beneficially combined with other methods. Without exceptions, combining DACS improves the performance of uncertainty-based methods by suggesting diverse samples as useful candidates. The improved performance is remarkable in RMNIST. For example, DACS increases the performance of LearnLoss, which shows a similar performance with Random, as much as or better than the diversity-based methods. This improvement could be attributed to preventing uncertainty-based methods from selecting redundant samples. In CIFAR-10 as well, the largest performances are achieved when combining DACS with the uncertainty-based methods.

\subsection{Prediction of Drug-Protein Interaction}
\subsubsection{Dataset configuration.}
For the regression task, we perform a drug-protein interaction task, which is the task to predict the affinity between the pair of drug and protein. We evaluate the performance on Davis \cite{davis2011comprehensive}, and it roughly contains 20,000 training and 5,000 test pairs with its affinity. We follow the same pre-processing scheme as \cite{ozturk2018deepdta} and evaluate each method by mean squared errors. 

The active learning starts with randomly selected 1,000 labeled samples with 19,000 unlabeled samples. In each cycle, each method selects 1,000 samples from the unlabeled pool. We repeat the acquisition in ten times

\subsubsection{Models.}
We employ DeepDTA \cite{ozturk2018deepdta} as a backbone, which consists of two different CNNs for drug and protein. The concatenated vectors from each CNNs are fed to the fully-connected networks to predict the affinity. The parameters are optimized through the MSE loss. We train the networks using Adam \cite{kingma2014adam} optimizer with 0.001 learning rate for 50 epochs and set the mini-batch size to 128.

\subsubsection{Results.}
The comparison results are shown in Table \ref{exp:davis}. Here, we did not compare with NCE-Net and CDAL because they are optimized for the classification task. Performance trends between different methods are similar to the classification experiment. Again, DACS shows superior performance compared to other methods. The large accuracy margin between DACS and other methods in the initial cycles is remarkable and, in the last cycle as well, DACS shows approximately 11\% better performance compared to Core-set and LearnLoss. In addition, the performance of LearnLoss is largely increased when combined with DACS. Comprehensive results clearly reveal the strength of the proposed method not only in classification but also in the regression task.

\section{Analysis}
In this analysis, we answer important questions: $i)$ Does AL methods still work well with the small budgets? $ii)$ Is the sampling from sparse region indeed more effective than sampling from the dense one? $iii)$ Why does the selected subset from DACS lead to superior performance?

\subsection{Active Learning with Small Budget}
As stated in earlier, we follow the experimental settings of previous works \citep{yoo2019learning,agarwal2020contextual,wana2021nearest}. This settings typically have the size of query (i.e., $b$) at least more than 1,000 samples. However, there are possible scenarios that the labeling costs are highly expensive. In this case, we are only capable to query the small number of samples to oracles. To confirm the strength of DACS in such settings, we conduct the same experiments with the main experiments, but reduce the query size. For CIFAR10 and RMNIST, we use the same initial labeled dataset but set the query size to 100. For Davis, we query 500 samples in each cycle of active learning. The other settings are same with that of the main experiments. The results are shown in Figure \ref{fig:small_exp}. Similar to the main experiments, DACS shows superior performance over the entire tasks. Specifically, the remarkable performance gap between DACS and others is observed in RMNIST where redundant samples exist. These results verify that DACS still works quite well in the small number of query settings.

\begin{figure}[t]
\centering 
\subfloat[CIFAR-10]{%
  \includegraphics[width=0.8\linewidth]{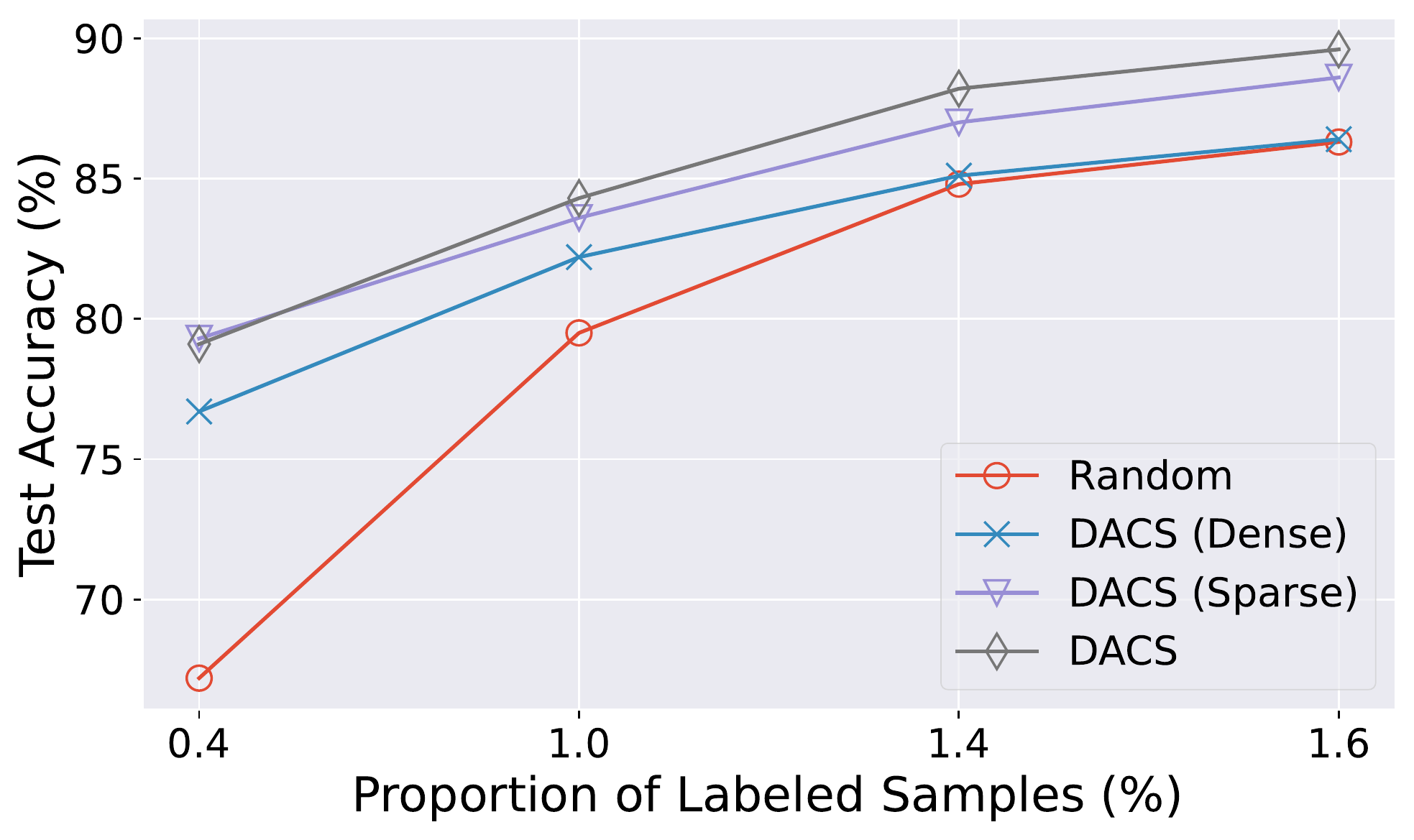}%
  \label{exp:global}
}

\subfloat[RMNIST]{%
  \includegraphics[width=0.8\linewidth]{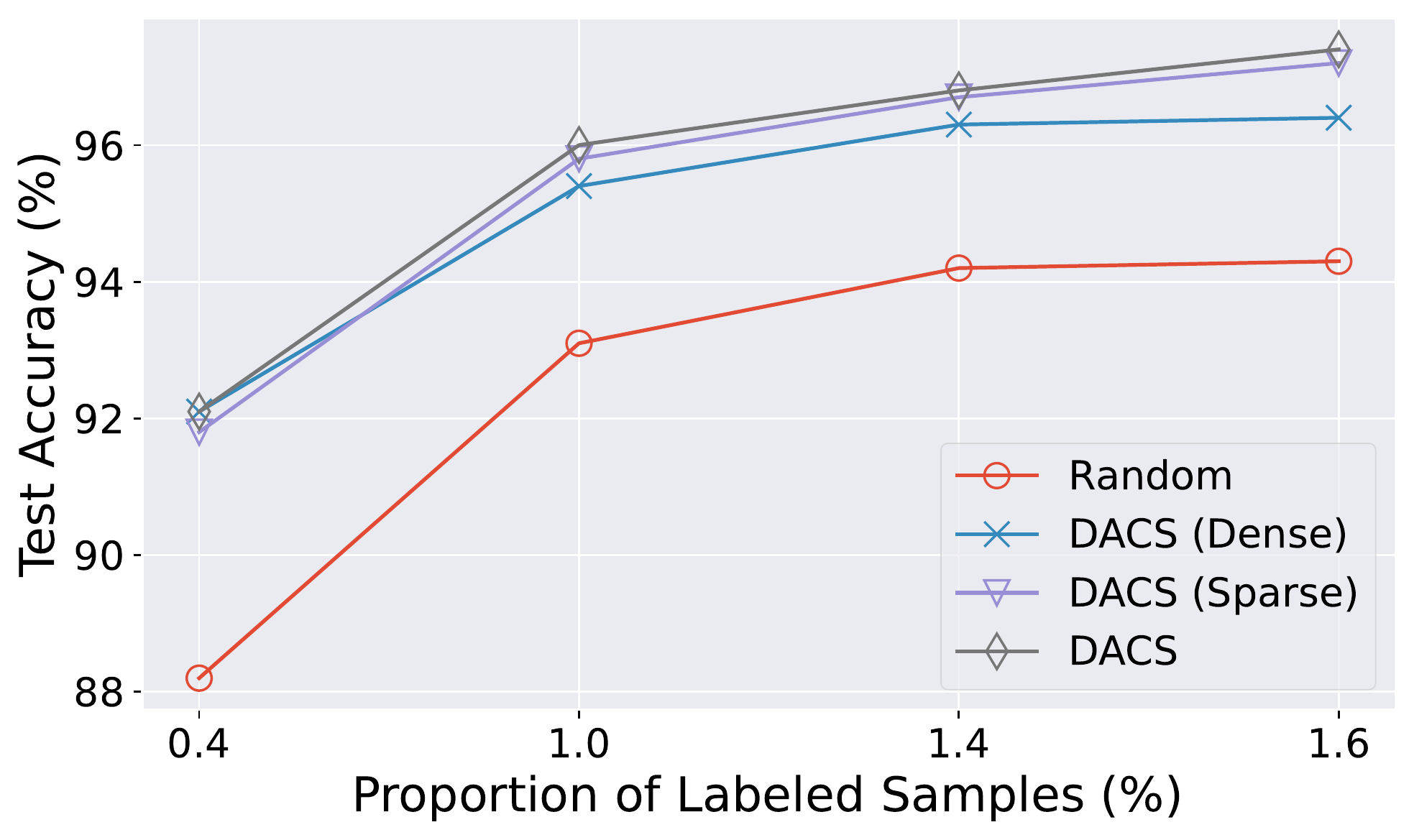}%
  \label{exp:local_drfit}
}
\caption{Performance of different sampling strategies. This shows that sampling from the sparse region is more effective than sampling from the dense.}
\label{fig:ablation}
\end{figure}

\subsection{Effectiveness of Dense and Sparse region}  

To answer the second question, we compare the performance when selecting samples only from the dense or sparse region. Here, unlabeled samples are split into three clusters (i.e., $h = 3$ in Eq. \ref{cluster}) based on the estimated density, and we measure the performance of sampling from the most dense and sparse clusters except for the intermediate cluster. Experimental settings are the same as CIFAR-10 and the results are shown in Table \ref{fig:ablation}. We can see that sampling from the sparse region results in better performance than sampling from the dense region. A noticeable point is that the performance of the dense region is gradually on par with the Random method, indicating that sampling from the dense region gradually fails to select informative samples compared to sampling from the sparse region. The results also present that DACS, which utilizes multiple acquisitions depending on the density, performs better than the single acquisition (i.e., sampling only from sparse or dense).

\subsection{Subset Diversity and Information Analysis}\label{just}




Diversity-based methods consider sample diversity, and uncertainty-based methods take into account informativeness. We quantitatively analyze the selected samples from the different methods based on what each method considers to answer the last question. Similar to \cite{wana2021nearest}, we quantify the subset informativeness as the normalized event probability by following information theory \cite{mackay2003information} and define the diversity using the average distance between selected samples. 

\begin{figure}[t]
\centering 
\subfloat[Subset Information]{%
  \includegraphics[width=0.8\linewidth]{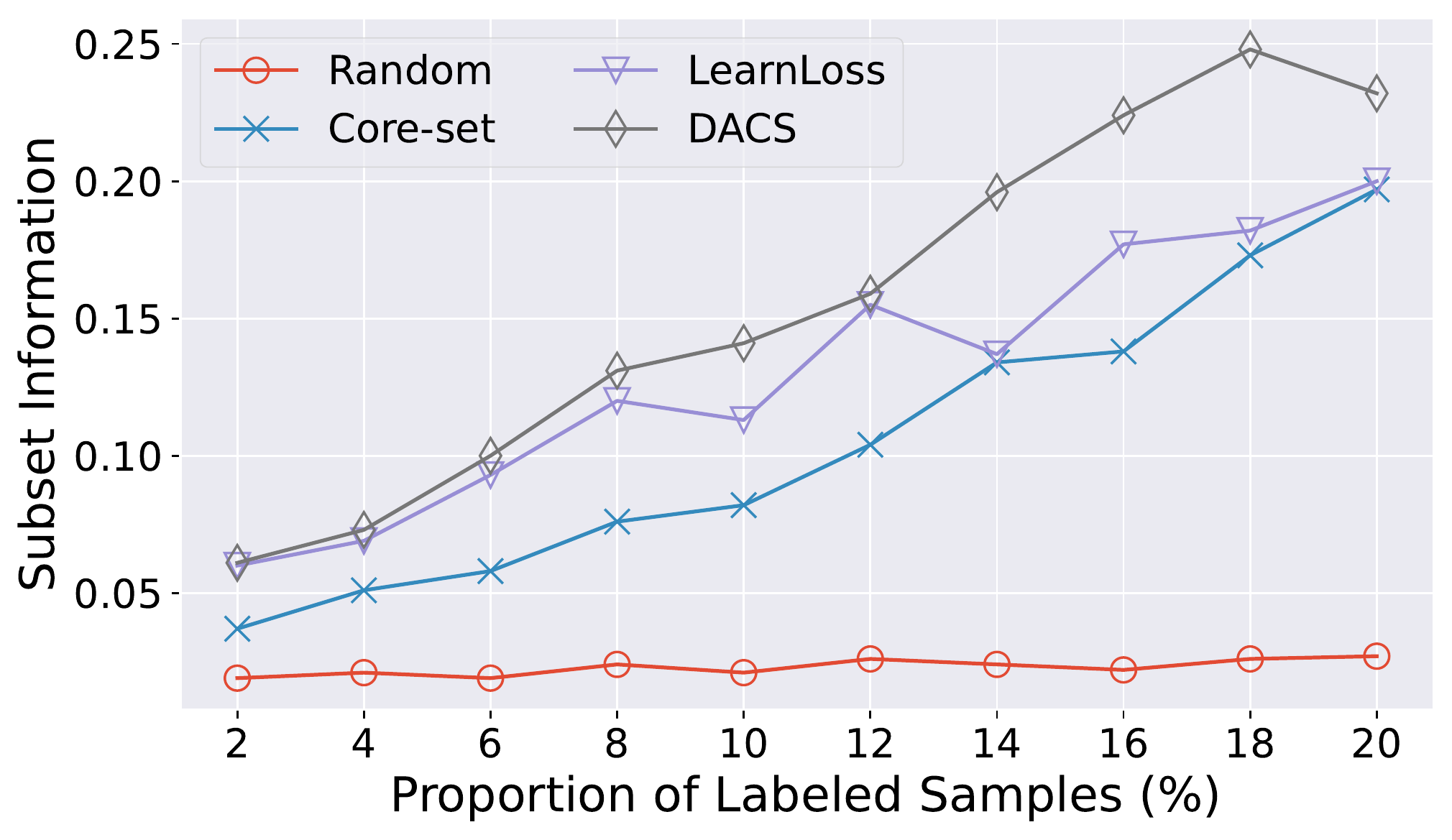}%
  \label{exp:global}
}

\subfloat[Diversity]{%
  \includegraphics[width=0.8\linewidth]{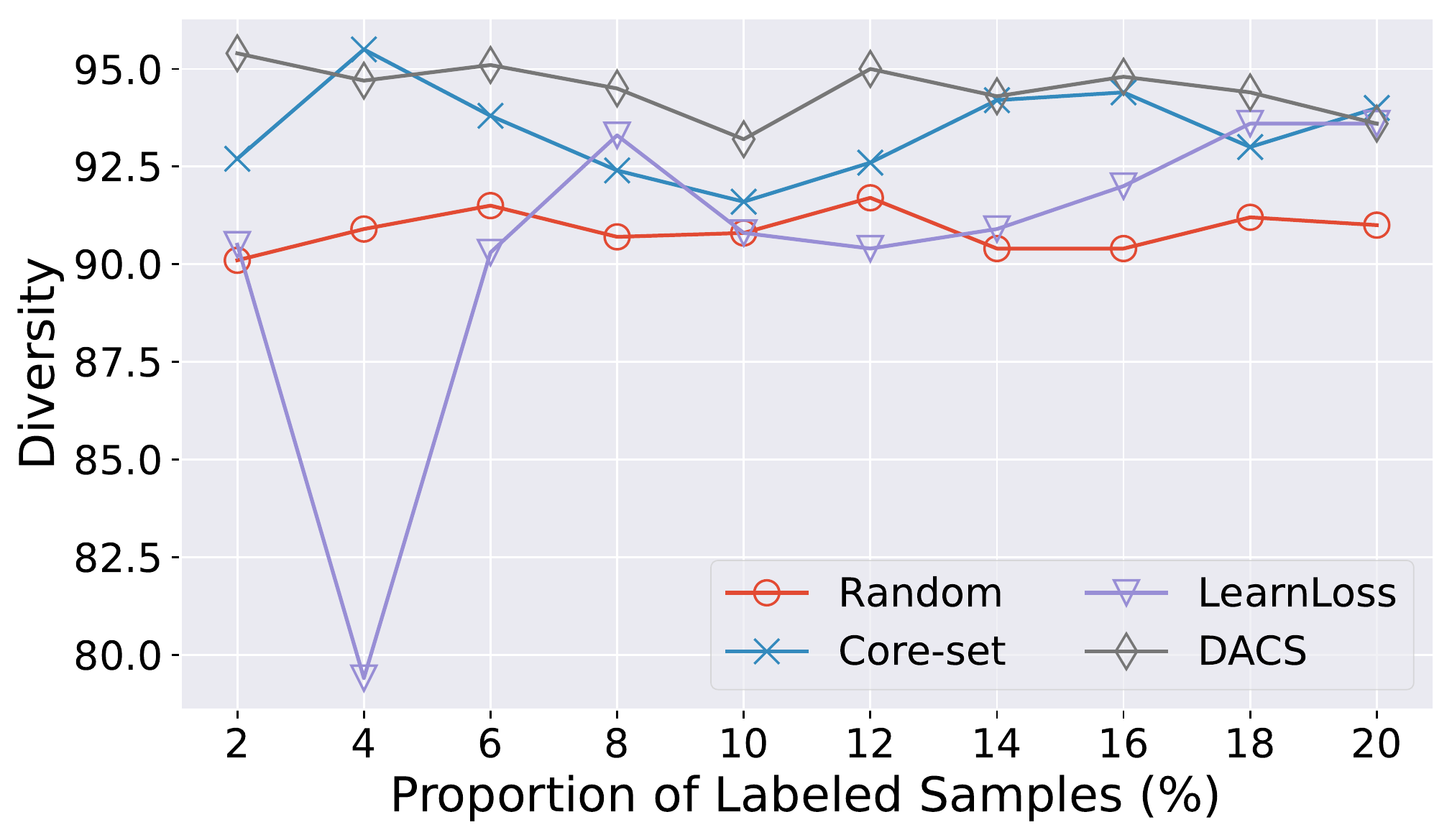}%
  \label{exp:local_drfit}
}
\caption{Subset information (Left) and diversity (Right) of selected samples in ten cycles of the active learning. This indicates that DACS can take advantages from both informativeness and diversity.}
\label{anal:subset_div}
\end{figure}

Based on the two measures, we evaluate the subset selected from diversity-based method (Core-set), uncertainty-based method (LearnLoss), Random, and DACS. We use the experimental settings of CIFAR-10 and the results are shown in Figure \ref{anal:subset_div}. Understandably, the selected samples from LearnLoss show higher informativeness than Core-set as the former explicitly considers the informativeness. When it comes to the diversity, Core-set exhibits its strength over the LearnLoss. Compared to these baselines, the selected samples from DACS show superior quality in both metrics. Particularly, the informativeness result (Figure \ref{anal:subset_div} (Left)) indicates that the DACS selects informative samples fairly well although informativeness has not been explicitly considered in the process. These results not only justify the effectiveness of the proposed method but show that DACS could take the strength from both the diversity- and uncertainty-based methods by empowering the core-set to be aware of the density of feature space.

\section{Related works}\label{sec:related_work}

The uncertainty-based methods can be categorized according to the definition of uncertainty. In the beginning, the posterior probability of the predicted class is popularly used as an uncertainty measure \citep{lewis1994heterogeneous,lewis1994sequential}, and these are generalized to the prediction entropy \citep{settles2008analysis,joshi2009multi,luo2013latent}. Recently, various definitions have been proposed to mainly solve classification tasks. For example, Yoo et al. \cite{yoo2019learning} train a loss prediction model, and the output of which serve as an uncertainty surrogate. Sinha et al. \cite{sinha2019variational} and Zhang et al. \cite{zhang2020state} learn the feature dissimilarity between labeled and unlabeled samples by adversarial training, and they select the samples having most dissimilar to the labeled ones. Different from these works, Wan et al. \cite{wana2021nearest} define the rejection or confusion on nearest neighbor samples by replacing the softmax layer with the prototypical classifier. Bayesian approaches \citep{gal2017deep,kirsch2019batchbald} are also proposed, however, they suffer from the inefficient inference and the convergence problem.

Diversity-based methods select samples that are most representative of the unlabeled data. Among these methods, clustering-based methods are frequently used in the literature \citep{nguyen2004active,dasgupta2008hierarchical}. Huang et al. \cite{huang2010active} extend the strategy of clustering methods by combining the uncertainty, but it is only applicable to the binary classification. Yang et al. \cite{yang2015multi} maximize the diversity by imposing a sample diversity constraint on the objective function. Similarly, Guo et al. \cite{guo2010active} performs the matrix partitioning over mutual information between labeled and unlabeled samples. However, it is infeasible to apply the above two methods to large unlabeled datasets since they requires inversion of a very large matrix (i.e., $|\mathcal{U}_t| \times |\mathcal{U}_t|$). Sener et al. \cite{sener2017active} solve sample selection problem by core-set selection and show promising results with a theoretical analysis. Agarwal et al. \cite{agarwal2020contextual} extend the idea to capture semantic diversity by estimating the difference in probability distribution between samples.

A few studies have considered the density in AL \citep{zhu2009active,xu2007incorporating}. However, these methods utilize the density as a secondary method for the uncertainty-based method, and they even do not use it for the diverse sampling. More importantly, these works prefer dense regions, which includes a number of highly similar samples, unlike DACS that primarily exploits sparse regions.




\section{Conclusion}
This paper has proposed the density-aware core-set (DACS) method which significantly improves the core-set method with the power of the density-awareness. To this end, we have analyzed the feature space through the lens of the local density and, interestingly, observed that the samples in locally sparse regions are highly informative than the samples in dense regions. Motivated by this, we empower the core-set method to be aware of the local density. DACS efficiently estimate the density of the unlabeled samples and divide the all feature space by considering the density. Afterward, the samples in the sparse regions are favorably selected by decomposed selection algorithm of the core-set. The extensive experiments clearly demonstrate the strength of the proposed method and show that DACS can produce a state-of-the-art performance in the real-world setting where redundant samples exist. We believe that our research can help in environments that require expensive labeling costs such as drug discovery \cite{shin2019self,beck2020predicting}.

\bibliographystyle{ACM-Reference-Format}
\bibliography{sample-base}










\end{document}